\documentclass{article}

\usepackage[nonatbib]{neurips_2024}
\nolinenumbers

\usepackage[utf8]{inputenc} % allow utf-8 input
\usepackage[T1]{fontenc}    % use 8-bit T1 fonts
\usepackage{hyperref}       % hyperlinks
\usepackage{url}            % simple URL typesetting
\usepackage{booktabs}       % professional-quality tables
\usepackage{amsfonts}       % blackboard math symbols
\usepackage{nicefrac}       % compact symbols for 1/2, etc.
\usepackage{microtype}      % microtypography
\usepackage{graphicx}
\usepackage{amsmath} 
\usepackage{cleveref}
\usepackage{multirow}
\usepackage{graphicx}
\newcommand{\tabref}[1]{Table~\ref{#1}}
\newcommand{\figureref}[1]{Figure~\ref{#1}}

\title{ACE Metric: Advection and Convection Evaluation for Accurate Weather Forecasting}

\author{%
  Doyi Kim, \hspace{3mm} Minseok Seo, \hspace{3mm}Yeji Choi
  \\
  SI Analytics\\
  70, Yuseong-daero 1689beon-gil, Yuseong-gu, Daejeon, Republic of Korea \\
  \texttt{\{doyikim,minseok.seo,yejichoi\}@si-analytics.ai} \\
}

\begin{document}

\maketitle

\begin{abstract}
Recently, data-driven weather forecasting methods have received significant attention for surpassing the RMSE performance of traditional NWP (Numerical Weather Prediction)-based methods.
However, data-driven models are tuned to minimize the loss between forecasted data and ground truths, often using pixel-wise loss.
%However, data-driven models are tuned to minimize the loss between input data and ground truths, often using MSE loss, comparing pixel-based differences.
%
This can lead to models that produce blurred outputs, which, despite being significantly different in detail from the actual weather conditions, still demonstrate low RMSE values.
%This can lead to models that produce blurred outputs, which, despite losing detailed information, still demonstrate low RMSE values.
%
Although evaluation metrics from the computer vision field, such as PSNR, SSIM, and FVD, can be used, they are not entirely suitable for weather variables.
This is because weather variables exhibit continuous physical changes over time and lack the distinct boundaries of objects typically seen in computer vision images.
%
%While evaluation metrics from the computer vision field, such as PSNR, SSIM, and FVD, could be employed, they are not entirely suitable for weather variables, which possess characteristics vastly different from typical computer vision images.
%
To resolve these issues, we propose the advection and convection Error (ACE) metric, specifically designed to assess how well models predict advection and convection, which are significant atmospheric transfer methods.
We have validated the ACE evaluation metric on the WeatherBench2 and MovingMNIST datasets.
\end{abstract}

\section{Introduction}

%
%Weather forecasting, based on massive data and numerical equations, rapidly absorbs deep learning approaches.
%
Data-driven weather forecasting rapidly advances in various directions, including weather-specific model architectures~\cite{bi2022pangu, lam2022graphcast, ravuri2021skilful, park2023long}, objective functions~\cite{andrychowicz2306deep}, and data augmentation methods~\cite{seo2022domain}.
%
%DL-based weather forecasting models~\cite{bi2022pangu,lam2022graphcast, pathak2022fourcastnet,ravuri2021skilful,andrychowicz2306deep} have already been proposed for various scales and lead times in this context.
%
These data-driven models are being utilized for various scale scenarios, from medium-range global forecasts~\cite{bi2022pangu,pathak2022fourcastnet,lam2022graphcast} of weather variables such as temperature and wind fields, to short-term regional precipitation~\cite{andrychowicz2306deep,ravuri2021skilful} forecasts.
%
%Hundreds of times, faster computation speeds reduce computational costs and enable rapid responses to hazardous weather.
%
%Data-driven approaches, compared to traditional numerical weather prediction (NWP) models that use supercomputers to formulate the physical rules of atmospheric states into partial differential equations (PDEs) and solve them using numerical simulations~\cite{dee2011era}, can make real-time predictions even on a single GPU~\cite{pathak2022fourcastnet}, establishing a new direction in the field of weather forecasting.
Unlike traditional numerical weather prediction (NWP) models, which use supercomputers to formulate physical rules of atmospheric states into partial differential equations (PDEs) and solve them using numerical simulations~\cite{dee2011era}, data-driven approaches can make real-time predictions even on a single GPU~\cite{pathak2022fourcastnet}. They are establishing a new direction in the field of weather forecasting.
%
%Also, the performance of recently demonstrated data-driven models proves they can be used as operating models.
These advantages are particularly beneficial for countries that cannot operate supercomputers or require rapid responses to hazardous weather.
Beyond computational efficiency, recent data-driven weather forecasting models have also achieved better RMSE values~\cite{lam2022graphcast} in medium-range forecasts than operational NWP models like integrated forecast system (IFS).
%

\begin{figure*}[t!]
 \centering
 \includegraphics[width=1.0\linewidth]{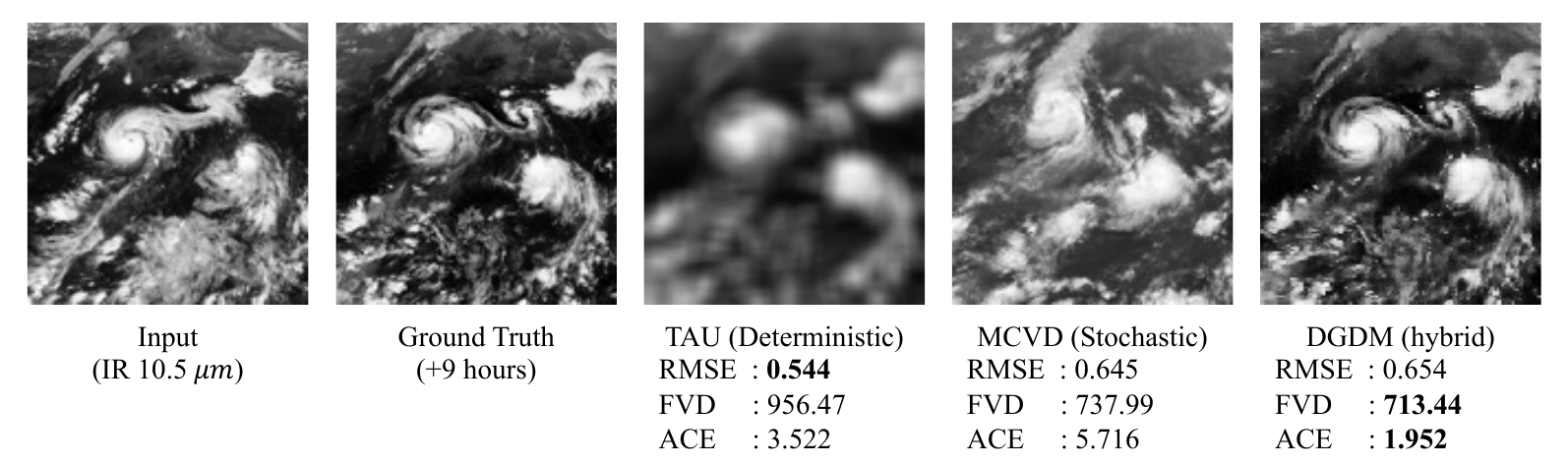}
 \caption{Comparison of RMSE, FVD, and ACE metrics in state-of-the-art deterministic, stochastic, and hybrid video prediction models.}
 \label{fig:teaser}
\end{figure*}

%Most models stem from computer vision fields, so data processing, model structure, and evaluation methods follow that rule.
%
%However, weather data have physical meaning for each pixel, different from RGB,  so it requires more considered processing.
%
%
In traditional weather forecasting with NWP, RMSE is usually used to measure model performance.
However, is it appropriate to use RMSE also for evaluating data-driven weather forecasts?
In typical deep learning tasks like image generation or video prediction~\cite{tan2022simvp,tan2023temporal,shi2015convolutional,srivastava2015unsupervised}, data-driven models trained with MSE often produce blurry outputs that achieve low RMSE values but do not align well with human perception~\cite{zhang2018unreasonable, unterthiner2018towards}.
To address this issue, metrics such as FVD~\cite{unterthiner2018towards}, LPIPS~\cite{zhang2018unreasonable}, and FID~\cite{heusel2017gans} have been proposed and are widely used in the field of image generation.
Unfortunately, these methods are also not suitable for evaluating data-driven weather forecasts.
~\cref{fig:teaser} shows samples of RMSE and FVD measurements for state-of-the-art methods based on observational data.
The figure shows that the blurriest image obtained the best RMSE skill score. At the same time, FVD measured how perceptual the image was (how well the pixel distributions matched), regardless of how well the model predicted the actual weather.
Furthermore, as shown in ~\cref{tab:tab_comprision}, common deep-learning metrics like MAE, MSE, PSNR, SSIM, and LPIPS are also unsuitable for evaluating weather forecasting.
Relying on a single evaluation metric to assess and improve data-driven weather forecasting models may not be helpful in practical applications.

To solve this problem, we propose an advection and convection error (ACE) metric that can simultaneously evaluate the horizontal movement and vertical development of weather variables.
% %Advection and convection, the horizontal and vertical motions of fluids, can explain meteorological phenomena at various scales.
Advection and convection, representing the horizontal and vertical motions of fluids, are crucial for understanding meteorological phenomena at various scales.
% Generally, large-scale atmospheric motions caused by temperature and pressure change are mainly driven by advection, and the smaller-scale convection terms can be ignored in comparison.
Generally, large-scale atmospheric motions caused by temperature and pressure changes are primarily driven by advection, the horizontal movement of air masses.
%However, dangerous weather, such as heavy precipitation, is caused by deep convection, a phenomenon in which air rises rapidly.
Conversely, convection refers to the vertical movement of air, which is essential for phenomena such as thunderstorms and heavy precipitation.
% ------------------0521
%To alleviate this problem, we proposed wind-based data augmentation recipes for weather forecast models in Seo et al. (2022), and as the next step, we propose a new evaluation metric.
%
%However, it is not easy to capture vertical motion from 2D data.
%
%As a result, most data-driven models that use model data (e.g., ERA5) as input only show high accuracy for large-scale phenomena related to advection. However, extreme weather predictions still have room for improvement.
%In addition, micro- to mesoscale weather events (e.g., thunderstorms, MCC) mainly utilize high-resolution observation data from satellites and radars rather than numerical model data.
%
%Multiple observation data are combined for accurate prediction. Still, the observation data has distinctive geospatial information, which means that even with the same cloud system, the coordinates will vary slightly depending on the observation position and angle.
%
Convection occurs when warm air rises, cools, and condenses to form clouds and precipitation.
This process is vital for predicting hazardous weather events, such as deep convection leading to severe storms.
\begin{table}[h!]
\centering
\caption{Summary of Characteristics for Various Evaluation and ACE Metrics}
\label{tab:tab_comprision}
\resizebox{0.8\columnwidth}{!}{%
\begin{tabular}{l|ccccccc}
\hline \hline
\multicolumn{1}{c|}{\textbf{Characteristic}} & \textbf{MAE} & \textbf{MSE} & \textbf{PSNR} & \textbf{SSIM} & \textbf{FVD} & \textbf{LPIPS} & \textbf{ACE} \\ \hline
Higher values for blurred images  & \checkmark   & \checkmark   & \checkmark  &       &   &    &     \\
Specializes in human visual perception  &  & &  & \checkmark   &   &    &     \\
Not focused on prediction accuracy           &   &    &     &      & \checkmark     & \checkmark   &     \\
Sensitive to slight misalignments    & \checkmark    & \checkmark  & \checkmark    &   &     &                &     \\ \hline \hline
\end{tabular}%
}
\end{table}

Therefore, advection is crucial for global-scale weather forecasting as it helps to understand and predict large-scale weather patterns influenced by the distribution of heat and moisture.
On the other hand, convection is vital for regional weather forecasting, where accurate prediction of vertical air movements is necessary to anticipate localized severe weather events, such as thunderstorms and heavy rainfall.
Thus,  it is reasonable to consider both advection and convection to analyze weather forecasting performance.
By simultaneously evaluating the horizontal and vertical motions of weather variables, we can gain a more comprehensive understanding of model accuracy and performance.
This approach accurately predicts both large-scale patterns and localized severe weather events, ultimately improving the reliability and utility of weather forecasts in real-world applications.

%
%However, existing data-driven models use MSE and MAE as evaluation metrics, which are highly sensitive to weather forecasting using multi-data. Our proposed method, advection and convection error (ACE) is intended to understand the 3D atmospheric structure and evaluate model performance.

%
\section{Related Work}
This section comprehensively investigates data-driven weather forecasting methodologies and explains the public open benchmark datasets proposed for developing data-driven weather forecasting models.
Also, we provide a detailed discussion of the evaluation metrics used in the field of weather forecasting.
\subsection{Benchmark Dataset}
Weather forecasting tasks are broadly categorized into short-term, mid-term, and long-term forecasting.
Short-term forecasting involves high-resolution predictions over small areas, which is particularly challenging due to the numerous factors influencing local weather.
Veillette and Samsi \textit{et al.}~\cite{veillette2020sevir} introduced the SEVIR dataset to address these challenges by providing a large-scale, annotated collection of over 10,000 rain events.
Each event includes multi-type image data from satellites (GOES-16), weather radar (NEXRAD), and GLM (Geostationary Lightning Mapper) sensors aligned spatially and temporally.
This dataset facilitates the training and validation of deep learning models for tasks such as precipitation nowcasting and synthetic weather radar generation and includes evaluation metrics for performance assessment.
They selected evaluation metrics such as RMSE, MAE, and MSE to measure the performance of short-term precipitation forecasts and weather variable predictions.
However, MSE and RMSE could yield good scores even when the forecasting outputs are blurred, as these metrics improve when the predictions are averaged.

For global medium-range forecasts, the ECMWF Reanalysis v5 (ERA5)~\cite{dee2011era} dataset provided by ECMWF is commonly used.
ERA5 provides data at six-hour intervals, includes 13 pressure levels, and covers 62 meteorological variables, making it a comprehensive but complex dataset.
However, using ERA5 as a benchmark dataset is challenging because researchers need to download and preprocess the data individually. 
In addition, each model uses different periods and atmospheric variables for training.
To address these issues, the low-resolution WeatherBench~\cite{rasp2020weatherbench} dataset, representing about 550 km per pixel (approximately 5.5 degrees), was proposed.
However, considering practical application scenarios, a benchmark that utilizes ERA5 at full resolutions was needed.
Consequently, the WeatherBench2~\cite{rasp2023weatherbench} dataset was introduced, with a resolution of 1440x721, corresponding to 0.25 degrees per pixel (about 25 km).

Nevertheless, in global medium-range forecasting, metrics such as MSE and RMSE were used for evaluating and training models, and data-driven models often achieved lower MSE and RMSE than operational models by producing blurred predictions.

\subsection{Data-driven Weather Forecasting}
Data-driven weather forecasting models have evolved into regional and global models. Shi~\cite{shi2015convolutional} ~\textit{et al.} proposed the ConvLSTM, which combines convolution with LSTM to better capture spatiotemporal correlations and consistently outperforms FC-LSTM.
This model has been successfully applied to regional precipitation forecasting. Subsequently, Ravuri~\cite{ravuri2021skilful} \textit{et al.} successfully applied generative adversarial networks for short-term precipitation forecasting in the UK, while Andrychowicz~\cite{andrychowicz2306deep} \textit{et al.} effectively utilized transformer structures for successful very short-term precipitation forecasting across the United States.

However, metrics such as MSE, RMSE, and CSI are used to evaluate the performance of these networks. 
Yet, CSI is unsuitable for evaluating models as it considers the model entirely incorrect if the prediction is shifted by even a single pixel.
Furthermore, a common issue with CSI, MSE, and RMSE is that models can achieve good scores generally by making blurred predictions.

Data-driven global weather forecasting models have garnered significant attention recently for achieving higher performance with significantly less computational demand compared to traditional NWP (Numerical Weather Prediction) models, which rely on supercomputers.
Pathak~\cite{pathak2022fourcastnet} ~\textit{et al.} successfully employed adaptive Fourier neural operators~\cite{guibas2021adaptive} for global weather forecasting up to 10 days ahead.
Following this, Xie~\cite{bi2022pangu} ~\textit{et al.} surpassed the medium-term forecasting performance of NWP models using an earth-specific transformer block, while Lam~\cite{lam2022graphcast} ~\textit{et al.} advanced the state-of-the-art by utilizing GNNs with a multi-mesh message-passing structure. These remarkable advancements have proposed a new paradigm in weather forecasting.

However, the authors themselves have pointed out that these models tend to produce blurred predictions, highlighting a critical issue.
Therefore, developing and employing fair metrics that can accurately evaluate and address this blurring problem is essential.
\section{Method}
\subsection{Preliminaries}
In meteorology, the movement of variables such as heat, moisture, and air is analyzed through convection, which represents vertical movements, and advection, which represents horizontal movements.
Convection is key to understanding many weather events as it helps form clouds and severe weather, such as thunderstorms and hurricanes, enabling meteorologists to predict when and where these events will occur.
It also moves heat up through the atmosphere, influencing the development of clouds and weather fronts, and even can spread or gather air pollutants, thereby affecting air quality. The equation for convection is calculated as follows:

\begin{equation}
\frac{\partial T}{\partial t} = -w \frac{\partial T}{\partial z}
\end{equation}
This equation shows that the rate of temperature change over time \(\frac{\partial T}{\partial t}\) is related to the product of the vertical velocity \(w\) and the vertical temperature gradient \(\frac{\partial T}{\partial z}\).
However, in most situations, it is not possible to measure the exact height \(z\), so indirect indicators are used.
Variables such as heat, moisture, and air rise vertically as their temperature increases, which means their intensity increases when observed from a top-bottom view.
Assuming there is no horizontal movement, Equation (1) is redefined for scenarios with multi-time point data \(t1, t2\) as follows:

\begin{equation}
\Delta T = T(t2) - T(t1)
\label{eq:convection_modi}
\end{equation}

where \(\Delta T\) represents the temperature change observed between two consecutive time points, indicating possible convection activities when analyzed over time.

Advection moves heat and moisture across different regions, shaping weather by forming clouds and driving storms.
It also spreads air pollutants.
This process is crucial for forecasting the movement and transformation of weather systems.
It enhances preparedness for future meteorological conditions.
The advection equation can be expressed as:
\begin{equation}
\frac{\partial \phi}{\partial t} + \mathbf{u} \cdot \nabla \phi = 0
\label{eq:advection}
\end{equation}
Here, \( \phi \) represents the scalar field being advected (e.g., temperature or moisture), \( \mathbf{u} \) is the velocity vector of the fluid, and \( \nabla \phi \) is the spatial gradient of \( \phi \).

\textbf{Note on mingling Convection and Advection over Time $t$} Convection and advection are intricately intertwined due to the dynamic nature of the atmosphere.
These processes often occur simultaneously and influence each other, making it challenging to analyze them separately.
\begin{figure*}[t!]
 \centering
 \includegraphics[width=0.8\linewidth]{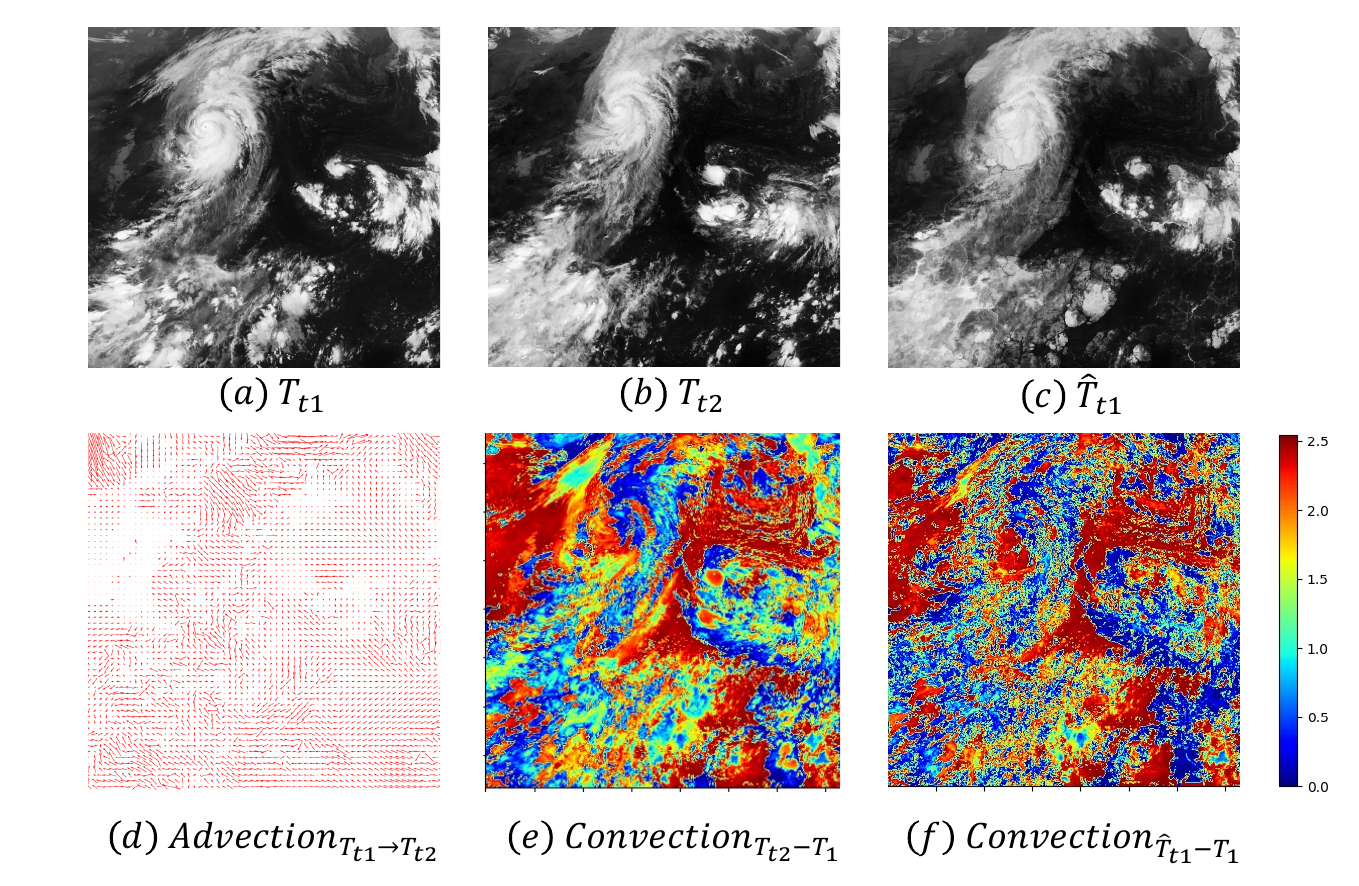}
 \caption{Figure estimating Convection and Advection in infrared (IR) images from geostationary weather observation satellites. (c) involves extracting advection and warping the $T_{t2}$ image to generate the t1 image, excluding convection. (d) represents advection between t1 and t2, while (e) shows the difference between t2 and t1, and (f) illustrates the difference between the estimated t1 and the original t1. \textbf{Note}: There is a 10-hour difference between $T_{t1}$ and $T_{t2}$.}
 \label{fig:vis4}
\end{figure*}
\subsection{Convection and Advection Error}
The Advection and Convection Error (ACE) metrics assess the accuracy of the weather forecasting model $f_{model}$. They evaluate how effectively the model, using the initial image $I_{o}$, predicts the horizontal movements (advection) and vertical development (convection) in the forecasted image $\hat{I}_{f}$, compared to the ground truth $I_{f}$.
Since both convection and advection simultaneously influence the changes between \(I_{o}\) and \(I_{f}\) over time, it is necessary to separate them. Therefore, we first measure the horizontal movement as advection, and then remove the extracted advection \(\Delta(v_{x},v_{y})\) in \(I_{f}\) to create \(\hat{I}_{o}\). Subsequently, we extract the convection between \(I_{o}\) and \(\hat{I}_{o}\) using ~\cref{eq:convection_modi}

\paragraph{Advection Error}
In \cref{eq:advection}, which calculates advection over time, the wind speed component $u$ is often not available or its use is restricted in most meteorological satellite imagery or NWP (Numerical Weather Prediction) data.
Therefore, an alternative method is necessary to estimate advection from every single meteorological variable.

In the image domain, under the assumption that intensity is conserved over time, the equation for calculating optical flow from an image is as follows:
\begin{equation}
\frac{\partial I}{\partial t} + \frac{\partial I}{\partial x} v_x + \frac{\partial I}{\partial y} v_y = 0,
\label{eq:optical}
\end{equation}
where \(I\) represents the image intensity at any given point, \(v_x\) and \(v_y\) are the horizontal and vertical components of the velocity field, respectively, indicating the rate of change of the image position in each direction.

Expressing \cref{eq:optical} in the form of advection, where \(\mathbf{v} \cdot \nabla I = v_x \frac{\partial I}{\partial x} + v_y \frac{\partial I}{\partial y}\), it can be rearranged as follows:
\begin{equation}
\mathbf{v} \cdot \nabla I + \frac{\partial I}{\partial t} = 0
\label{eq:adad2}
\end{equation}
This rearrangement emphasizes how the components of the velocity field \(v_x\) and \(v_y\) interact with the spatial derivatives of the image intensity to maintain the conservation of intensity over time.
Ultimately, this highlights that the equation is essentially equivalent to the advection equation.
However, since actual meteorological variables, including convection and energy, are not perfectly conserved over time, it is impossible to directly find a solution that satisfies ~\cref{eq:adad2}.
Therefore, it is necessary to extract the advection velocity field through numerical optimization-based techniques.
The following equation is the numerical optimization-based formula used to extract the velocity field in the ACE (Advection and Convection Error) analysis:

\begin{equation}
  \begin{minipage}{0.7\linewidth}
    \includegraphics[width=\linewidth]{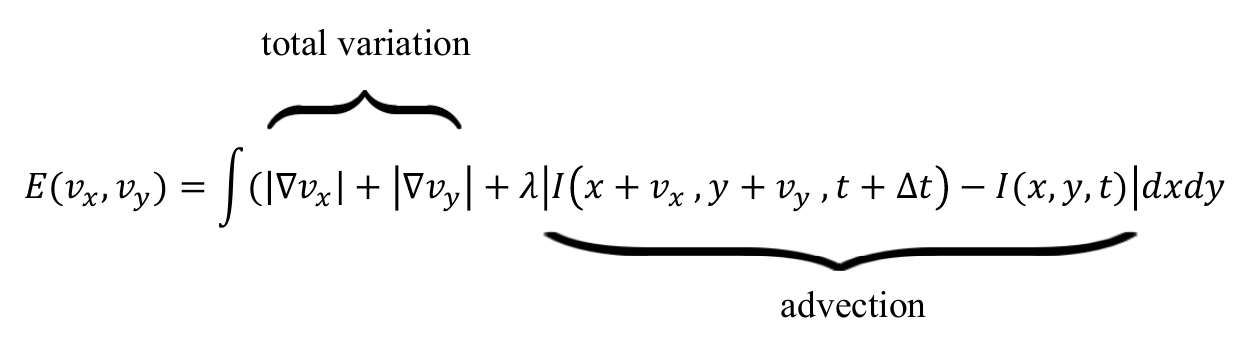} %
  \end{minipage}
  \label{eq:tvl1}
\end{equation}
Thus, the final advection error, when the function \(f_v\) is used to extract the velocity field by minimizing the above equation, follows the formula below:
\begin{equation}
AE = | f_{v}(I_{o}, I_{f}) - f_{v}(I_{o}, \hat{I}_{f}) | 
\end{equation}
In ~\cref{eq:tvl1}, the total variation term penalizes large variations between \(I_{o}\) and \(\hat{I}_{f}\). Therefore, if \(\hat{I}_{f}\) is more blurred compared to \(I_{o}\), the AE error will increase.

\paragraph{Convection Error}
By reversing the advection determined between \(I_{o}\) and \(I_{f}\) using \(f_{v}\) on \(I_{f}\), the transformation results in \(\hat{I}_{o}\), which retains only the changes in intensity from \(I_{o}\) over time with all advection effects removed.
Therefore, when the remapping function is denoted as \( f_{\text{re}} \), the Convection Error is defined as follows:
\begin{equation}
CE = \left|  | I_{o} - f_{\text{re}}(\hat{I}_{f}, \hat{v}_{x}, \hat{v}_{y}) | - \left| I_{o} - f_{\text{re}}(I_{f}, v_{x}, v_{y}) \right| \right|
\end{equation}
Thus, the final ACE metrics are defined as follows:
\begin{equation}
ACE = AE + \frac{CE}{AE}
\end{equation}
When AE is large, indicating significant advection errors, the term \(\frac{CE}{AE}\) becomes smaller, reducing the relative impact of the convection error (CE).
This is appropriate because when advection errors are large, the reliability of the convection error assessment is diminished.
As a result, it is reasonable to lessen the influence of CE within the overall ACE computation.
This adjustment ensures that, despite potentially significant advection errors, the contribution of convection errors is appropriately moderated, maintaining the integrity of the overall ACE assessment.

\section{Experiments}
\paragraph{Datasets} We selected the WeatherBench2~\cite{rasp2023weatherbench}, a global meteorological variable forecasting benchmark dataset, MovingMNIST~\cite{srivastava2015unsupervised}, a synthetic dataset for predicting the movement of two digits, and SEVIR~\cite{veillette2020sevir}, a regional precipitation prediction dataset, to evaluate the ACE metrics.
\begin{itemize}
    \item \textbf{WeatherBench2 .} The WeatherBench2 dataset is designed for benchmarking machine learning models in medium-range (1-14 day) global weather forecasting at 6-hour intervals.
    The dataset covers the period from 1959 to 2023 and includes various weather-related variables, featuring 62 variables across 13 pressure levels.
    It also provides high-resolution spatial data, with grid points spaced at approximately 0.25 degrees globally.

    We selected data from the year 2020 to validate the ACE evaluation metrics.
    We chose state-of-the-art data-driven weather forecasting models for comparative analysis, including PanguWeather~\cite{bi2022pangu} and Graphcastnet~\cite{lam2023learning}.
    Additionally, we selected the IFS model for numerical weather prediction (NWP).
    \item \textbf{MovingMNIST.} The Moving MNIST dataset consists of 10,000 video sequences, each containing 20 frames, where two digits move independently within a 64 x 64 pixel frame.
    The digits, randomly selected from the training set, begin at random initial positions. Each digit is assigned a velocity with a direction uniformly chosen randomly on a unit circle and a magnitude selected from a predetermined range.
    Throughout the sequences, the digits frequently intersect and overlap when they occupy the same space, and they bounce off the frame of edges.
    This synthetic dataset is ideal for evaluating our metrics as it ensures the conservation of energy.

    We conducted a comprehensive evaluation of our ACE metrics across various models, ranging from those developed in 2015 to the latest in 2023~\cite{shi2015convolutional, lotter2016deep, wang2022predrnn, wang2019memory, yu2020efficient, guen2020disentangling, chang2021mau, gao2022simvp, tan2023temporal, tan2022simvp}, to ensure their validation across different model architectures.
\end{itemize}
\paragraph{Evaluation metrics.} To compare ACE with other common metrics, we selected Mean Absolute Error (MAE), Root Mean Squared Error (RMSE), Mean Squared Error (MSE), Peak Signal-to-Noise Ratio (PSNR), Structural Similarity Index (SSIM), and Frechet Video Distance (FVD).
MAE, RMSE, and MSE are widely utilized in various fields, including statistics and machine learning, particularly for regression problems or models where the accuracy of continuous variable predictions is crucial.
PSNR, SSIM, and FVD are commonly used in image processing and video analysis to evaluate the quality of outputs relative to a reference.

\paragraph{Implementation details.} We utilized a numerical optimization method with the following hyperparameters: \(\tau = 0.25\), \(\lambda = 0.15\), \(nscales = 5\), \(warps = 5\), \(\epsilon = 0.01\), \(innerIterations = 30\), \(outerIterations = 10\), \(scaleStep = 0.8\), and \(medianFiltering = 5\).
The parameter \(\tau\) controls the update step size, while \(\lambda\) balances the data fidelity and regularization terms.
The multi-scale pyramid processing, controlled by \(nscales\) and \(scaleStep\), allows capturing motion at different levels of detail.
Warping steps per scale, determined by \(warps\), aid in aligning images more accurately. The convergence threshold \(\epsilon\) ensures iterations stop when changes between successive estimates are minimal.
Inner and outer iterations, set by \(innerIterations\) and \(outerIterations\) respectively, manage the optimization process.
Median filtering with a kernel size of 5 was applied to reduce noise in the resulting flow.
These settings were chosen to optimize the balance between computational efficiency and accuracy in the flow estimation process.

\subsection{MovingMNIST Results}
~\tabref{tab:tab2} presents the quantitative comparison of the ACE metric with other metrics in the Moving MNIST dataset.
As illustrated in ~\tabref{tab:tab2}, the ACE metric demonstrates similar trends to other metrics such as MSE, MAE, SSIM, PSNR, and FVD in the Moving MNIST dataset.
Notably, methods that explicitly consider advection and convection, such as TAU, MAU, and PhyDNet, exhibit higher ACE scores.
These experimental results indirectly suggest that the ACE metric effectively accounts for advection and convection phenomena.

\begin{figure*}[t!]
 \centering
 \includegraphics[width=1.0\linewidth]{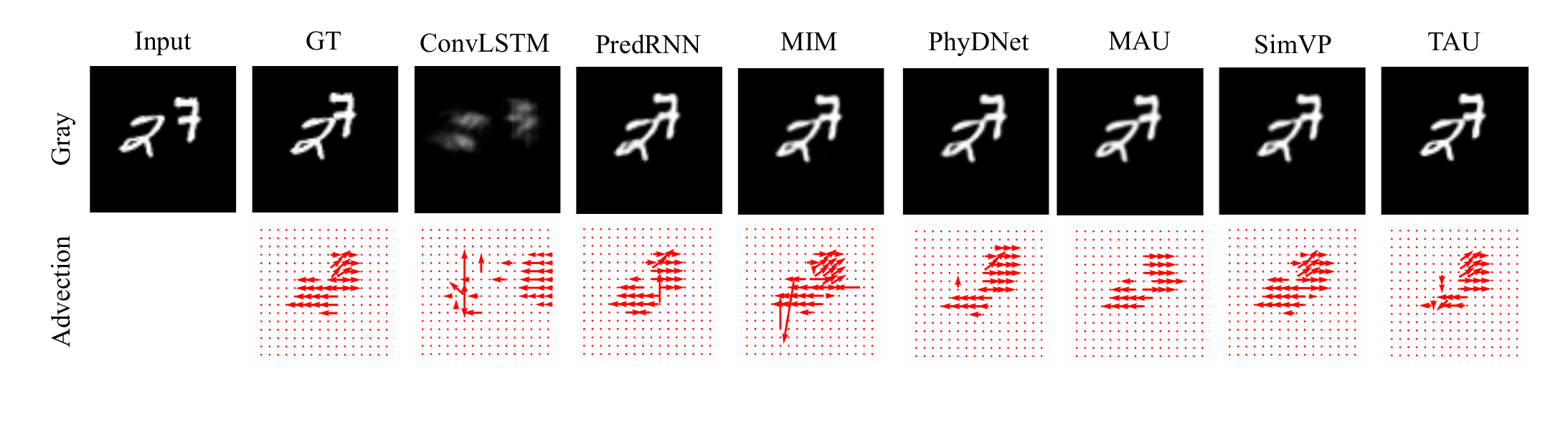}
 \caption{An illustration of advection visualized on the Moving MNIST dataset. The arrow direction represents t2 to t1. Bold indicates the highest score and underline indicates the second highest score.}
 \label{fig:vis44}
\end{figure*}

~\figureref{fig:vis44} visualizes the results of the baseline models. Even though the visual differences among each baseline result seem minimal, the ACE metric, which takes into account the total variation when extracting advection, shows a significant difference in advection. This clear distinction is expected to be more helpful in developing our models.

%
%
% 0.33 & 0.13 & 0.72 MIM
% 0.53 & 20.31 & 0.68 phydnet
% 0.33 & 0.07 & 0.68 MAU
% 0.33 & 0.13 & 0.72 Simvp
% 0.22 & 0.07 & 0.55 tau

\begin{table}[t!]
\renewcommand{\arraystretch}{1.5} 
\centering
\caption{Comparison table of video forecasting methods on the Moving MNIST dataset, measured using traditional performance metrics and the ACE metric.}
\label{tab:tab2}
\resizebox{0.8\columnwidth}{!}{%
\begin{tabular}{c|ccccc|ccc}
\hline \hline
\multirow{2}{*}{\textbf{Method}} & \multirow{2}{*}{\textbf{MSE}} & \multirow{2}{*}{\textbf{MAE}} & \multirow{2}{*}{\textbf{SSIM}} & \multirow{2}{*}{\textbf{PSNR}} & \multirow{2}{*}{\textbf{FVD}} & \multicolumn{3}{c}{\textbf{ACE}} \\ \cline{7-9} 
&  &  &  &  &  & AE   & CE   & $AE+\frac{CE}{AE}$  \\ \hline
ConvLSTM~\cite{shi2015convolutional}    &  29.80               & 90.64     &  0.9288 & 22.10 &  79.19 &   1.97 & 0.23 & 2.09  \\ 
PredRNN~\cite{lotter2016deep}           &  25.14               & 77.85     & 0.9283  & 22.53 &  50.40 &  0.35  & 0.19   & 0.90   \\ 
PredRNN++~\cite{wang2022predrnn}        &  \underline{23.97}   & 72.82     & \underline{0.9462} & \underline{23.28}&  45.73 &   \underline{0.26} & 0.16 & 0.87      \\ 
MIM~\cite{wang2019memory}               &  \textbf{22.55}      & \textbf{69.97}  &\textbf{0.9498} & \textbf{23.56} &  47.53 &    0.33 & 0.13 & 0.72    \\ 
PhyDNet~\cite{guen2020disentangling}    &  28.19               & 78.64      & 0.9374 & 22.62 &  38.75 &    0.53 & 20.31 & 0.68    \\ 
MAU~\cite{chang2021mau}                 &  26.86               & 78.22      & 0.9398 &22.57 & \underline{36.09}  &   0.33 & \underline{0.07} & \underline{0.6}       \\ 
SimVP~\cite{gao2022simvp}               &  32.15               & 89.05      & 0.9268 &21.84  & 72.96   &   0.33 & 0.13 & 0.72           \\ 
TAU~\cite{tan2023temporal}              &  24.60               & \underline{71.93}  & 0.9454 & 23.19 &   \textbf{28.16} &  \textbf{0.22} & \textbf{0.07} & \textbf{0.55}           
          \\ \hline \hline
\end{tabular}%
}
\end{table}

\subsection{WeatherBench2 Results}
\begin{figure*}[t!]
 \centering
 \includegraphics[width=1.0\linewidth]{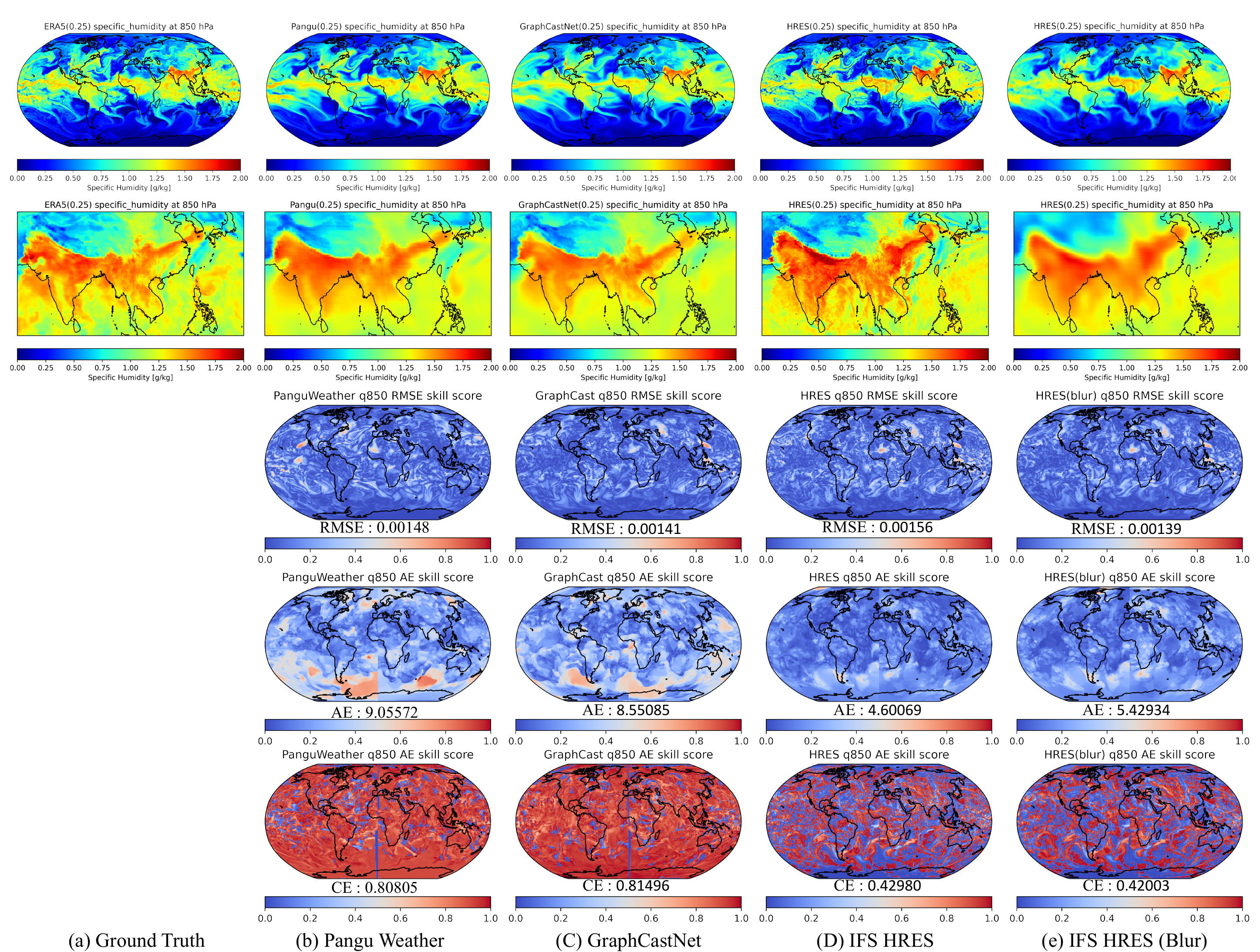}
 \caption{The first row shows the prediction results and ground truth for a +5 days forecast. The second row presents a zoomed-in view of a specific region. The third row displays the RMSE map, the fourth row shows the AC map, and the fifth row depicts the CE map.r}
 \label{fig:vis6}
\end{figure*}

\paragraph{Did data-driven models actually outperform the NWP model?} Most data-driven weather forecasting models are evaluated using the RMSE value.
However, is it really true that a higher RMSE value indicates better performance?
~\figureref{fig:vis6} visualizes the results of the data-driven models PanguWeather and GraphCastNet, the NWP model IFS HRES, and a blurred version of IFS HRES using a (15x15) kernel.
As shown in the figure, PanguWeather and GraphCastNet appear more blurred than IFS HRES, with IFS HRES (15x15) appearing the most blurred.
The RMSE values are 0.00148, 0.00141, 0.00156, and 0.00139, respectively.
These experimental results indicate that RMSE favors models that produce blurrier outputs.
So, can we truly call IFS HRES (15x15) state-of-the-art?
To answer this question, we need a new metric that can better interpret data-driven weather forecasting models.

The fourth and fifth rows of ~\figureref{fig:vis6} visualize AE and CE, respectively.
As shown in the figure, advection, which is relatively easier to predict, is reasonably well-forecasted by both data-driven models and the NWP model.
Conversely, convection, known to be more challenging to predict, shows higher errors across all models.
Quantitatively, the ACE scores for PanguWeather, GraphCastNet, IFS HRES, and IFS HRES (15x15) are 9.144, 8.646, 4.694, and 5.506, respectively.
These experimental results demonstrate that the ACE metric does not improve with artificial blurring; instead, it decreases.
This indicates that the ACE metric robustly evaluates model performance even in blurring.
Moreover, according to the ACE metric, data-driven models, despite showing significant potential, still perform lower than the NWP models.

\paragraph{Why are blurred predictions limited in practical applications?}
Paris, with an area of 105.4 $km^{2}$, would be represented by approximately 4 pixels in a global weather forecasting model with a resolution of 0.25 degrees per pixel. Similarly, New York, which covers an area of 783.8 $km^{2}$, would be represented by about 31 pixels. Because of this low resolution, blurred predictions fail to provide the necessary accuracy for regional forecasts.

While such models may achieve high scores on a global scale, they are not suitable for city-level weather forecasting, where precision is crucial. Therefore, for regional weather forecasting, we need to produce sharp predictions. In particular, to accurately forecast regions heavily influenced by atmospheric flow and development, it is essential to effectively capture both convection and advection.

\begin{figure*}[t!]
 \centering
 \includegraphics[width=0.9\linewidth]{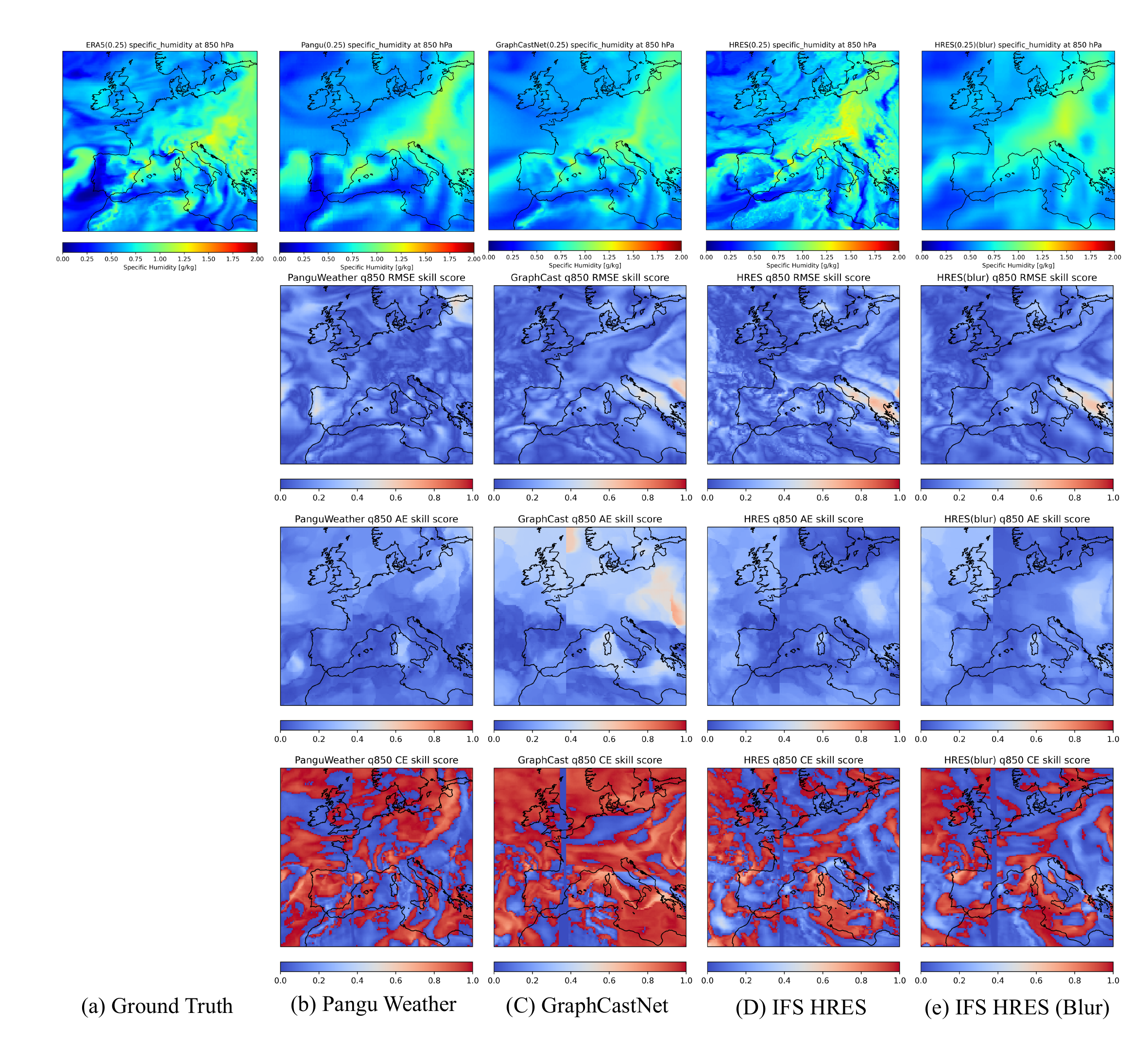}
 \caption{The first row shows the prediction results and ground truth for a +10 days forecast. The second row displays the RMSE map, the third row shows the AC map, and the fourth row depicts the CE map.}
 \label{fig:vis7}
\end{figure*}

As shown in ~\figureref{fig:vis7}, since each pixel covers 25 $km^{2}$, data-driven models that produce blurred predictions have an average error across all regions. In contrast, the sharp predictions made by the IFS HRES model may be significantly wrong in some areas but are notably accurate in others.
%

%\paragraph{Why is it important to evaluate advection?}
%
%Data-driven global weather models predict at 6-hour intervals for up to 10 to 14 days. But does pixel-by-pixel comparison truly aid in the development of weather forecasting models?
%
%For example, a pixel-by-pixel comparison can show a significant error if a single pixel is shifted after 10 days of forecasting.
%
%However, if only one pixel shifts after a 10-day forecast, it actually indicates that the model has performed quite well.
%
%Therefore, it is more appropriate to include the evaluation of horizontal movement (advection) when assessing weather forecasting models. 
%
\section{Conclusion}
In this paper, we introduced the Advection and Convection Error (ACE) metric to address the shortcomings of traditional evaluation metrics in data-driven weather forecasting.
Although metrics like RMSE, MSE, PSNR, and SSIM are widely used in deep learning, they often fail to accurately reflect the true performance of weather forecasting models, particularly in capturing the intricate patterns of advection and convection crucial for predicting weather phenomena.
The ACE metric specifically evaluates how well models predict the horizontal and vertical movements in weather data, providing a more comprehensive assessment of model performance.
By incorporating the dynamics of advection and convection, ACE effectively addresses the issue of blurred outputs often produced by models optimized with conventional loss functions.
Our experiments on the WeatherBench2 and MovingMNIST datasets demonstrate that ACE provides a more nuanced evaluation than traditional metrics.
The results show that methods designed to consider advection and convection explicitly achieve higher ACE scores, confirming the effectiveness of the metric in capturing these critical weather processes.
Moving forward, the ACE metric can be a robust tool for evaluating and improving data-driven weather forecasting models.
By offering clearer insights into how models predict complex atmospheric dynamics, ACE has the potential to enhance the reliability and accuracy of weather forecasts, ultimately contributing to better preparedness for severe weather events.
\clearpage
{\small
\bibliographystyle{plain}
\bibliography{egbib}
}

\end{document}